%% file: VA-Judger_arxiv.tex
\documentclass{article} % For LaTeX2e
\usepackage{iclr2027_conference,times}

\input{math_commands.tex}

\usepackage{hyperref}
\usepackage{url}
\usepackage{graphicx}
\usepackage{booktabs}
\usepackage{colortbl}
\usepackage{placeins}
\definecolor{bestresult}{RGB}{226,244,220}

\title{VA-Judger: Reward Modeling from Human Preference Feedback for Joint Video-Audio Generation}
\author{
\begin{minipage}{\dimexpr\textwidth-2\tabcolsep-3pt\relax}
\centering\normalfont
\vspace*{8pt}
{\bfseries
Yinming Huang\textsuperscript{1,2,*} \quad
Shuyuan Tu\textsuperscript{1,*} \quad
Xi Yan\textsuperscript{1} \quad
Zihan Yang\textsuperscript{1} \quad
Jianhua Han\textsuperscript{3} \\
Hang Xu\textsuperscript{3} \quad
Kaihang Pan\textsuperscript{4} \quad
Yu-Gang Jiang\textsuperscript{1,\textdagger} \quad
Zuxuan Wu\textsuperscript{1,2,\textdagger}} \\[3pt]
\textsuperscript{1}Fudan University \quad
\textsuperscript{2}Shanghai Innovation Institution \\[3pt]
\textsuperscript{3}Yinwang Intelligent Technology Co., Ltd \quad
\textsuperscript{4}Zhejiang University \\[3pt]
\textsuperscript{*}Equal contribution. \quad
\textsuperscript{\textdagger}Corresponding authors. \\[3pt]
Code: \href{https://github.com/ShareLab-SII/VA-Judger}{\texttt{https://github.com/ShareLab-SII/VA-Judger}}
\end{minipage}
}

\iclrfinalcopy
\begin{document}

\maketitle

\begin{figure}[h!]
    \centering
    \includegraphics[width=\linewidth]{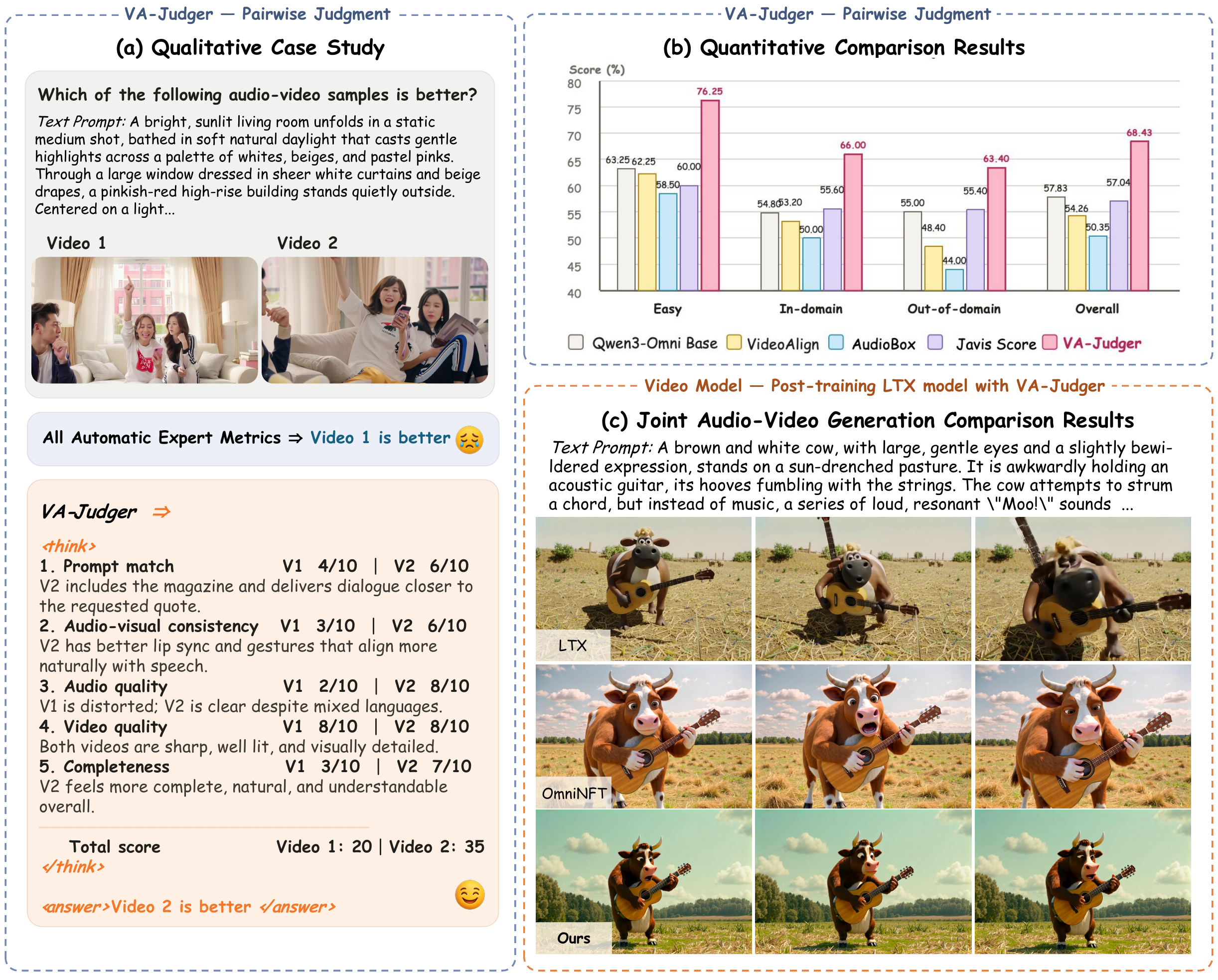}
    \caption{Evaluation of VA-Judger. (a) VA-Judger selects the preferred clip through structured dimension-wise reasoning, whereas separate evaluation metrics disagree with human judgment. (b) VA-Judger achieves the highest agreement with human preferences. (c) VA-Judger provides the reward for post-training LTX-2, improving the quality over the base model and OmniNFT.}
    \label{fig:teaser}
\end{figure}

\begin{abstract}
Using reinforcement learning to post-train joint video-audio generation models requires a reward signal. Existing methods construct this reward by combining metrics for individual quality dimensions, including audio quality, visual fidelity, and synchronization.
However, these metrics evaluate perceptual dimensions separately and fail to capture the overall semantic and temporal coherence among the text prompt, video, and audio that shapes human preferences. 
More critically, they are poorly aligned with actual human judgments. Optimizing models against these metrics encourages reward hacking, generating video-audio content that achieves high scores on these metrics yet appears incoherent or unfaithful to human viewers. 
To address this problem, we first construct \textbf{VAPref-10K} for joint video-audio generation, comprising 9K prompts and 10.3K fine-grained paired comparisons from generation models. We also introduce the \textbf{VA-Judger-Bench} benchmark with both in-domain and out-of-domain model comparisons to evaluate whether reward models truly align with human preferences. 
We further propose \textbf{VA-Judger}, a chain-of-thought omni-reward model for joint video-audio generation. 
In particular, VA-Judger first learns from pairs with clear quality gaps to establish structured output and coarse preference discrimination, then distills reliable preference explanations for harder near-quality comparisons via rejection sampling verified against human annotations, and finally performs dimension-wise reinforcement learning that decomposes human feedback into individual quality dimensions for denser reward signals than a single binary preference label. 
Experiments show that VA-Judger outperforms metric baselines in predicting human preferences on both in-domain and out-of-domain evaluations. Using its human-aligned rewards to post-train an audio-video generation model also improves most automatic metrics and human preference.

\end{abstract}

\section{Introduction}
Recent advancements in joint video-audio generation~\citep{openai2025sora2, google2025veo31, seedance2026seedance, wanteam2026wan27, google2026geminiomni, wang2025universe, low2025ovi, liu2026javisdit, team2025kling, team2026mova, liu2026javisdit++, chern2026speed, hacohen2026ltx, zhou2026avgen} have made it possible to generate increasingly realistic visual content and synchronized sound. 
These inspire researchers to explore post-training via reinforcement learning to align the synthesized content with perceptual quality and user intent. 
However, the reward signal used during reinforcement learning remains a critical bottleneck for joint video-audio generation. Current approaches such as OmniNFT~\citep{zhang2026omninft} construct rewards from multiple evaluation metrics, including VideoAlign~\citep{liu2026improving} and HPSv3~\citep{ma2025hpsv3} for video quality, Audiobox Aesthetics~\citep{tjandra2025meta} for audio quality, CLAP~\citep{wu2023large} for audio-text alignment, and DeSync~\citep{iashin2024synchformer} for audio-visual synchronization.
These metrics score perceptual dimensions separately and may miss semantic and temporal relations among the prompt, video, and audio.
More fundamentally, such evaluation metrics are poorly aligned with actual human judgments~\citep{liu2026improving,wang2024lift}.    
A sample may receive a high synchronization score while depicting the wrong event, obtain a strong visual score while carrying emotionally incompatible sound, or satisfy several metric criteria while failing as a coherent video-audio experience (Figure~\ref{fig:teaser}(a,b)). Optimizing models against such rewards encourages overfitting to narrow metric targets, synthesizing video-audio content that scores well on these metrics yet appears unfaithful to human viewers, undermining the practical value of post-training (Figure~\ref{fig:teaser}(c)). 

This issue has been addressed in other generation domains through human preference learning. In language modeling, preference-based optimization and reward modeling have enabled instruction-following generation~\citep{christiano2017deep, ouyang2022training, rafailov2023direct}. Image generation has adopted human-preference datasets and reward models such as Pick-a-Pic~\citep{kirstain2023pick}, ImageReward~\citep{xu2023imagereward}, HPSv3~\citep{ma2025hpsv3}, and MPS~\citep{zhang2024learning}. 
Video generation has similarly begun to leverage human feedback through InstructVideo~\citep{yuan2024instructvideo}, LiFT~\citep{wang2024lift}, VADER~\citep{prabhudesai2024video}, and VideoReward~\citep{liu2026improving}. 
However, none of these efforts address joint video-audio generation, and their reward models cannot be directly transferred to this setting. 
The fundamental challenge is that human preference over video-audio content is not decomposable into independent per-modality judgments. Text-to-image/video reward models do not observe the audio stream and thus cannot assess whether sound matches the depicted event, while audio-only evaluation lacks the visual context that shapes human perception of sound plausibility. 
Even combining such single-modality evaluators cannot recover the cross-modal interactions that govern human preference. 
Addressing this problem requires both joint audio-video assessment and preference data derived from holistic human judgments of the complete text-video-audio experience, neither of which currently exists for joint video-audio generation.

In light of this, we first address the scarcity of human preference data by constructing \textbf{VAPref-10K}, a large-scale preference dataset tailored for joint video-audio generation. VAPref-10K comprises approximately 9K prompts derived from captioned web videos across seven broad video-audio categories, with 10.3K fine-grained paired comparisons generated by state-of-the-art generation models.
Rather than absolute scalar scoring for each quality dimension, which can vary substantially across human experts due to subjective calibration, annotators compare two generated clips from the same prompt, select the preferred one, and identify the quality dimensions that justify their decision. 
From VAPref-10K, we introduce \textbf{VA-Judger-Bench} as a benchmark for evaluating whether reward models align with human video-audio preferences. Beyond in-domain evaluation, VA-Judger-Bench includes an out-of-domain split built from AVGenBench~\citep{zhou2026avgen}, pairing outputs from closed-source models (Kling~2.6~\citep{kuaishou2025kling26}, Wan~2.6~\citep{alibabacloud2026wan26}, Veo~3.1 Fast, Veo~3.1 Quality~\citep{google2025veo31}, and Sora~2~\citep{openai2025sora2}).
They are excluded from training.

Furthermore, we propose \textbf{VA-Judger}, a chain-of-thought omni-reward model for joint video-audio generation. Video-audio preference judgment is a complex reasoning task because it must simultaneously acquire a structured comparison format, discriminate subtle cross-modal failures, and provide dimension-level feedback. 
Thus, we adopt a progressive training strategy. We begin with pairs that have clear quality gaps, for which Gemini 3.1~\citep{google2026gemini31pro} generates rubric-format CoT answers containing dimension-wise scores, explanations, and a final preference to perform cold start. These generated answers help the model acquire the structured comparison format and basic quality discrimination before encountering ambiguous cases. 
For the more challenging comparisons between outputs with subtle quality differences, Gemini's final preferences alone are not sufficiently reliable. 
We thus use rejection sampling. Gemini 3.1~\citep{google2026gemini31pro} first generates CoT answers for hard pairs, and we retain only format-valid responses whose final preferences and human-selected dimension score orderings agree with the annotations.
This allows the model to learn from consistently formatted reasoning traces grounded in human judgments.
Finally, since a single binary preference label is too sparse to guide fine-grained preference learning, we perform dimension-wise reinforcement learning with GRPO~\citep{shao2024deepseekmath}. This separately rewards score orderings on human-identified quality dimensions, including visual quality, audio quality, synchronization, and semantic faithfulness, and therefore provides denser optimization signals than the final preference alone.

In conclusion, our contributions are as follows:
(1) We construct VAPref-10K and VA-Judger-Bench, comprising 10.3K human-labeled comparisons across three training models and five additional benchmark models.
(2) We propose VA-Judger, a chain-of-thought omni-reward model for joint video-audio generation, trained via easy-to-hard preference learning, structured-response distillation, and dimension-wise reinforcement learning with human feedback. To our knowledge, we are the first to explore reward modeling for joint video-audio generation via human feedback.
(3) We introduce VA-Judger-Bench for evaluating video-audio reward models against human preferences. VA-Judger achieves better alignment with human judgments in both in-domain and out-of-domain settings. Using VA-Judger as a reward for post-training further improves human-perceived generation quality.

\section{Related work}
\label{sec:related}

\subsection{Joint video-audio generation models}
Recent generative models have moved from silent video generation toward joint video-audio generation. Broader video-generation research explores unified free-form composition and reasoning in OmniWeaving and agentic long-form generation in ViMax~\citep{pan2026omniweaving,huang2026vimax}. Closed-source models~\citep{openai2025sora2, google2025veo31} demonstrate the joint-generation trend at product scale.
Regarding open-source models, UniVerse-1 stitches pretrained video and music experts~\citep{wang2025universe}, whereas Ovi uses twin diffusion backbones with blockwise cross-modal fusion~\citep{low2025ovi}. MOVA~\citep{team2026mova} and daVinci-MagiHuman~\citep{chern2026speed} use large-scale video data.
LTX-2~\citep{hacohen2026ltx} aims to handle the capacity imbalance between
modalities. 
Baton~\citep{tu2026baton} introduces explicit semantic blueprints for joint video-audio generation. 
The Javis~\citep{liu2026javisdit, liu2026javisdit++} series further emphasizes synchronization.
JoyAI-Echo~\citep{li2026joyai} targets long-form generation.
Complementary work studies motion editing~\citep{tu2024motioneditor,tu2024motionfollower} and identity-preserving or long-form portrait animation~\citep{tu2025stableanimator,tu2025stableanimator++,tu2025stableavatar,tu2026flashportrait}.
However, existing models are still primarily optimized by model-side objectives or expert metrics, leaving substantial room for improving human-perceived quality and video-audio coherence.

\subsection{Multimodal reward models}
Reward modeling is a standard way to turn human preference data into optimization signals. 
Recent multimodal generation methods improve unified visual modeling through discrete diffusion timestep tokens, zero-shot demonstrative instruction tuning, and intrinsic 3D awareness~\citep{pan2025generative,li2024fine,qiu2026spatialfusion}.
In image generation,
Pick-a-Pic~\citep{kirstain2023pick} provides image preference data, and ImageReward~\citep{xu2023imagereward} learns an image reward model. In video generation, LiFT~\citep{wang2024lift} trains LiFT-Critic for preference
alignment, and VideoReward~\citep{liu2026improving} studies multi-dimensional human preference annotations. 
Furthermore, IXC-2.5-Reward~\citep{zang2025internlm} is trained on broad image and video preference data, while UnifiedReward-Think~\citep{wang2026unified} provides reward signals for both multimodal understanding and generation tasks. However, neither model is trained on joint text-video-audio preference comparisons or explicitly evaluates whether the text, video, and audio form a coherent event. VA-Judger instead targets human preference alignment for joint video-audio generation by jointly assessing visual and audio quality, semantic fidelity, and cross-modal consistency.

\subsection{Reinforcement learning for generative models}
Recent works adapt online reinforcement learning to diffusion models. FlowGRPO~\citep{liu2026flow} makes GRPO applicable to text-to-image flow models.
DanceGRPO~\citep{xue2025dancegrpo} designs a unified GRPO framework for text-to-image/video and image-to-video generation. DiffusionNFT~\citep{zheng2026diffusionnft} uses positive and negative generations to define an implicit policy-improvement direction. 
ArcFlow improves two-step text-to-image generation through high-precision nonlinear flow distillation~\citep{yang2026arcflow}. SelfTok and Janus-Pro-R1 further combine unified visual generation with reinforcement learning or supervised and reinforcement-learning stages~\citep{wang2025selftok,pan2026janus}.
In joint video-audio generation, OmniNFT~\citep{zhang2026omninft} applies online diffusion reinforcement learning, while related systems such as JavisDiT++~\citep{liu2026javisdit++} explore preference optimization through different formulations. OmniNFT's optimization still depends on separate evaluation metrics such as VideoAlign~\citep{liu2026improving}, HPSv3~\citep{ma2025hpsv3}, Audiobox Aesthetics~\citep{tjandra2025meta}, CLAP~\citep{wu2023large}, and DeSync~\citep{iashin2024synchformer}. This motivates our dimension-wise reinforcement learning with VA-Judger, which learns the reward signal from human preference annotations instead of optimizing only narrow expert metrics.

\section{Method}
\label{sec:method}

\begin{figure}[t]
      \centering
      \includegraphics[width=\linewidth,trim={120bp 230bp 40bp 200bp},clip]{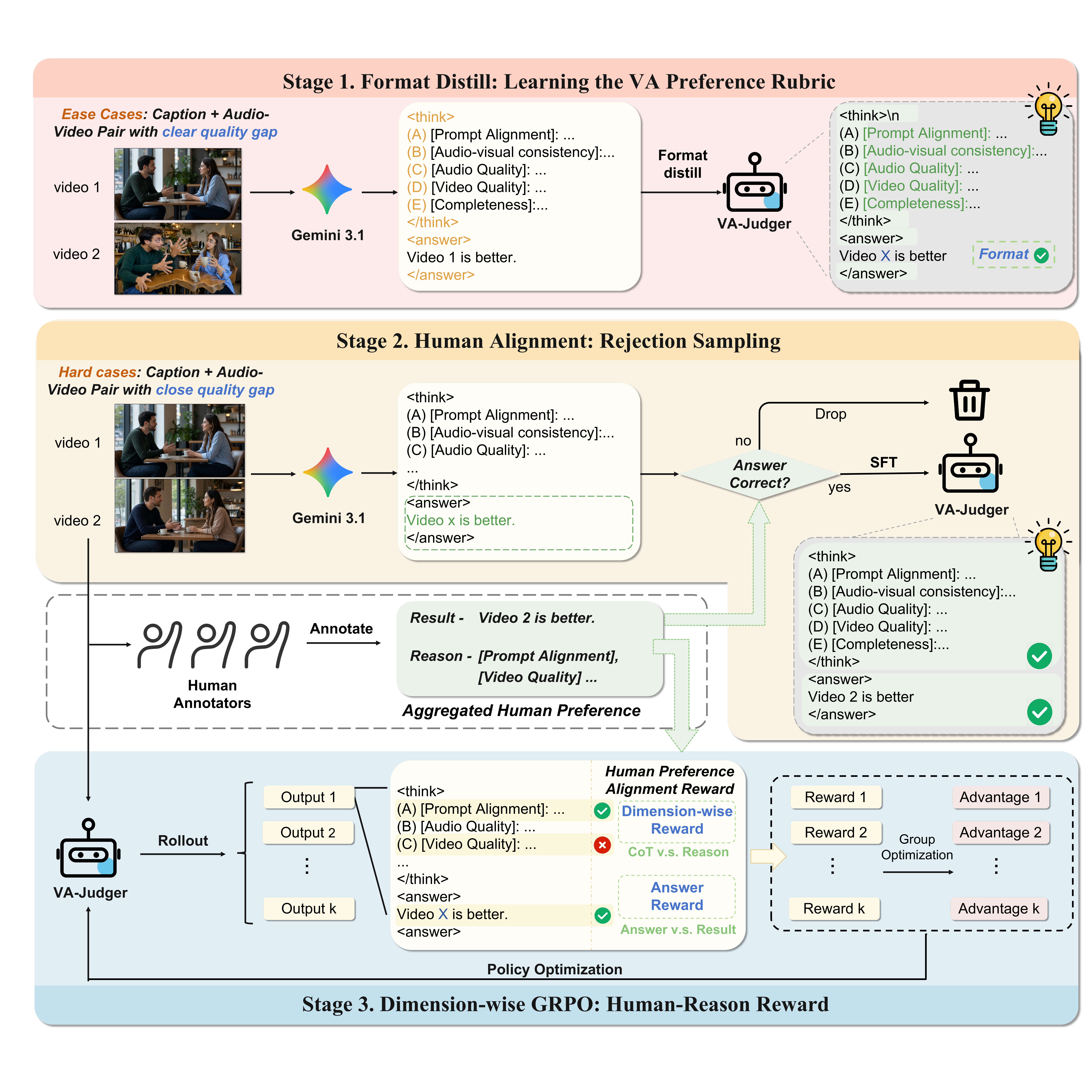}
      \caption{Overview of the VA-Judger training pipeline. Stage 1 cold-starts the model on easy preference pairs. Stage 2 aligns
      the model on harder pairs using responses filtered by final-choice agreement, dimension-score agreement, and format validation. Stage 3 performs GRPO using answer-level and dimension-level rewards grounded in human
      reasons.}
      \label{fig:architecture}
      \vspace{-0.4cm}
  \end{figure}

VA-Judger is designed to provide a reward signal for post-training text-to-video-audio generation models. 
Concretely, we first construct VAPref-10K from difficulty-aware video-audio pairs (Fig~\ref{fig:data_pipeline} and Sec~\ref{sec:data}). We then train VA-Judger using the three-stage pipeline (Fig~\ref{fig:architecture}), which combines an easy-pair cold start, hard-pair SFT using human-aligned and format-valid responses (Sec~\ref{sec:sft}), and dimension-wise reinforcement learning (Sec~\ref {sec:rl}).

\subsection{Task Formulation}
\label{sec:task}

We formulate video-audio reward modeling as pairwise preference prediction with structured dimension-wise reasoning.
Each example consists of a text prompt $p$, two generated video-audio clips $x_1=(v_1,a_1)$ and $x_2=(v_2,a_2)$, and a preferred index $y^\star \in \{1,2\}$. The input to the model is a fixed system prompt that specifies the evaluation rubric, followed by a user message containing the text prompt and the two videos with audio. 
This input format forces our model to jointly evaluate the combined text-video-audio performance, not a separate audio or video score.
The model then produces an interpretable chain-of-thought comparison~\citep{wei2022chain,wang2026unified}. 
For each clip, it assigns a 1--10 score with a brief justification on five dimensions, namely (A) prompt alignment, (B) video-audio consistency, (C) audio quality, (D) video quality, and (E) completeness and coherence, then aggregates these into a total score and emits a final preference in the format \texttt{<answer>video k is better</answer>}. 

\subsection{Human-Preference Data Construction}
\label{sec:data}

VAPref-10K contains human-annotated hard comparisons and out-of-domain comparisons. The 4,390 easy cross-model pairs form a separate cold-start set and are not part of VAPref-10K. 
To make VAPref-10K comprehensive, it covers a diverse range of text prompts across various scenarios. 
Recent benchmarks combine taxonomy-driven synthetic prompts with curated or video-derived examples~\citep{hua2026vabench,cao2025t2av,zhou2026avgen}. In contrast, VAPref-10K uses 10,173 real video-audio clips sourced from YouTube, Bilibili, films, and TV series as the primary source for prompt construction. This preserves naturally occurring events and cross-modal interactions, yielding more complex and realistic prompts. We trim the collected videos into 5-10-second segments and organize them into six categories.

As shown in Fig~\ref{fig:data_pipeline}, source videos are collected from the web and organized into six broad categories: multi-person dialogue, voice-over narration, music, cinematic ambience, environmental events, and talks. Source video distribution is shown in Fig~\ref{fig:video-category-distribution}. 
We then caption these source videos with Qwen3.5-Omni~\citep{team2026qwen3}. Although these captions capture rich multimodal detail, their temporally structured format produces dense timestamps and fragmented event descriptions in our data, making them unsuitable as direct inputs to video-audio generation models. 
Thus, we use a Qwen~\citep{yang2025qwen3} to remove the timestamp-heavy structure and rewrite each caption into a coherent, generation-ready \emph{text prompt} while preserving the visual content, audio events, speech, and cross-modal relations. This produces 10,083 rewritten prompt candidates. We then remove candidates that do not satisfy the generation and data-quality requirements. The remaining approximately 9K prompts are used to generate candidate clips and construct preference pairs.

Pair construction is deliberately difficulty-aware.
If all training pairs are near ties, the base omni-model may struggle to learn the rubric format and produce unstable judgments. If all pairs are easy, it may overfit to superficial model-specific cues and fail on realistic near-quality comparisons.
Thus, we build two complementary splits. 
The easy pool contains 4,390 pairs with clear quality gaps. Across all six categories, we compare Ovi~\citep{low2025ovi} with LTX-2~\citep{hacohen2026ltx}. In our preliminary evaluation, LTX-2 is substantially better than Ovi, so the preferred output is usually apparent from overall quality. For multi-person dialogue and voice-over, we also compare Ovi with daVinci-MagiHuman, which is strong in human-centric speech generation~\citep{chern2026speed}.
The raw hard pool contains 9,146 pairs with much smaller quality gaps. It compares two samples generated from the same prompt by the same model using different random seeds. Distinguishing these pairs requires close inspection of subtle audio-visual differences.
More details are in Sec~\ref{sec:sft}.

\subsection{Curriculum Supervised Fine-Tuning}
\label{sec:sft}

VA-Judger produces structured chain-of-thought judgments rather than opaque scalar scores. A base omni-model lacks this capability, as it is not trained to compare clips under a multi-dimensional rubric, and binary preference supervision alone cannot induce such structured reasoning.
Thus, we use supervised fine-tuning to establish the output format. 
We initialize VA-Judger from Qwen3-Omni~\citep{xu2025qwen3} and train it on complete rubric responses that include dimension-wise scores, explanations, total scores, and a final \texttt{<answer>} tag.

\subsubsection{Stage 1: Easy Cold-start}
\label{sec:sft-stage1}

Stage 1 aims to teach VA-Judger the required response format and lets the model learn preference discrimination from unambiguous examples. 
We begin with easy cross-model pairs, where large quality gaps make preferences clear and allow the model to learn comparison without relying on subtle perceptual differences. This easy-to-hard schedule follows curriculum learning, which improves optimization by presenting simpler examples before harder ones~\citep{bengio2009curriculum}.
We use Gemini 3.1 Pro~\citep{google2026gemini31pro} to generate structured reasoning responses, as it agrees with human annotators on 80\% of final preferences in a 200-pair pilot study, avoiding costly large-scale annotation at this stage. Training on these responses forces the model to follow the five-dimensional rubric, aggregate scores, and produce valid \texttt{<answer>} outputs, allowing the subsequent hard stage to focus on fine-grained preference alignment rather than output formatting.

\subsubsection{Stage 2: Hard Preference Alignment}
\label{sec:sft-stage2}

Hard pairs contain clips of similar overall quality, so their preference depends on subtle differences that must be identified through close inspection of the audio and video. This ambiguity makes the supervision generated by Gemini in Stage 1 unreliable. In a separate pilot study of 200 hard pairs, Gemini 3.1 Pro agrees with human judgments on only 60\% of the pairs. As the reward model is intended to represent subjective human preference~\citep{christiano2017deep,ouyang2022training}, we replace Gemini's final choices with human-provided preference labels. Annotators label all 9,146 raw hard pairs by selecting the preferred clip and identifying the rubric dimensions that support their decision.

Furthermore, human labels alone do not provide the complete structured responses required for SFT. We therefore use Gemini to generate rubric responses and retain only those that match the human preference, pass the regular-expression format parser, and assign the human-preferred clip a strictly higher score on every human-selected supporting dimension, that is, $s_{y^\star,d}>s_{3-y^\star,d}$ for all $d\in\mathcal D_{\mathrm h}$. This yields 4,436 hard-pair responses whose final choices and dimension-level score orderings agree with the human annotations. Training on these structured responses exposes VA-Judger to hard pairs with similar overall quality while preserving the response format learned in Stage 1.

\subsection{Dimension-wise Reinforcement Learning}
\label{sec:rl}
The two SFT stages teach the model to imitate structured rubric responses, but imitation alone has a fundamental limitation. The likelihood objective treats every token in the target response equally and does not distinguish whether the model's own generated judgments actually agree with human preferences. A model can achieve low SFT loss while still producing dimension scores that contradict its final answer, or choosing the wrong clip with confident but internally consistent reasoning. We therefore use an objective that rewards sampled judgments according to their agreement with human labels.

Thus, we propose \emph{Dimension-wise GRPO}, which extends GRPO~\citep{shao2024deepseekmath} with a human-grounded verifiable reward. A key observation is that a single binary reward (correct or incorrect final answer) is too sparse for structured judgments. The model may guess the right winner while assigning dimension scores that do not support its decision, learning a shortcut rather than dimension-level agreement. 
Our reward addresses this by jointly verifying two aspects of each response. Given input $x=(p,x_1,x_2)$ consisting of a prompt and two clips, the rollout policy samples $G$ responses $\{o_i\}_{i=1}^{G}$. 
Each human annotation provides the preferred clip $y^\star$ and a set of supporting dimensions $\mathcal D_{\mathrm h} \subseteq \mathcal D$ that justify the preference (the full rubric $\mathcal D$ is shown in Appendix Figure~\ref{fig:full-case}). For response $o_i$, we parse the predicted preference $\hat y_i$ and dimension scores $s_{i,k,d}$ for clip $k$ on dimension $d$. The reward combines an answer component that verifies the final preference with a dimension component that checks whether the score ordering on each human-specified dimension supports the human decision. For clarity, we define both components for a generic response $o$, with predicted preference $\hat y$ and score $s_{k,d}$ for clip $k$ on dimension $d$:
{\small
\begin{equation}
\label{eq:dimension_reward}
r_{\mathrm{ans}}(o) = \mathbf{1}[\hat{y}=y^\star],
\qquad
r_{\mathrm{dim}}(o) =
\frac{1}{|\mathcal{D}_{\mathrm{h}}|}\sum_{d\in \mathcal{D}_{\mathrm{h}}}
\mathbf{1}\!\left[s_{y^\star,d} > s_{3-y^\star,d}\right].
\end{equation}
}
For each sampled response, we define the combined reward as $R_i=\tfrac{1}{2}\bigl(r_{\mathrm{ans}}(o_i)+r_{\mathrm{dim}}(o_i)\bigr)$. This design rewards both the final choice and score orderings that agree with the human-identified quality dimensions. Responses whose dimension scores disagree with these dimensions receive only partial reward. Annotations without any valid dimension in $\mathcal D$, including Other-only annotations, are excluded from Dimension-wise GRPO, ensuring that $|\mathcal D_{\mathrm h}|>0$.

We optimize the policy with a group-normalized on-policy variant of GRPO. The $G$ rewards within each group are normalized to advantages $A_i$. We then maximize the advantage-weighted mean log likelihood of the generated tokens:
{\small
\begin{equation}
\label{eq:grpo_terms}
\begin{aligned}
A_i &= \frac{R_i-\bar R}{\sigma_R+\epsilon}, \qquad
\mathcal J_{\mathrm{GRPO}}(\theta) &= \mathbb E\!\left[
\frac{1}{G}\sum_{i=1}^{G}\frac{A_i}{|o_i|}
\sum_{t=1}^{|o_i|}\log \pi_\theta(o_{i,t}\mid x,o_{i,<t})\right].
\end{aligned}
\end{equation}
}
Here, $\bar R$ and $\sigma_R$ are the group mean and standard deviation of rewards, and $\epsilon$ stabilizes normalization. The submitted implementation uses neither PPO ratio clipping nor a reference-model KL penalty. This choice keeps the optimization objective consistent with the sampled group-relative advantages and avoids an additional reference-model copy.

\section{Experiments}
\label{sec:experiments}

\subsection{Experiment Setup}
\label{sec:experiment-setup}

\paragraph{Reward model evaluation.}
VA-Judger-Bench contains three splits: 400 easy pairs, 250 in-domain pairs, and 500 out-of-domain pairs. The easy split contains clear quality gaps and tests coarse preference discrimination. The in-domain split follows the hard training distribution and tests subtle judgments on familiar generation models. The out-of-domain split uses clips from unseen closed-source generation models to test the model's generalization capability. \textbf{None of the 1,150 benchmark pairs overlaps with the training set.}

\paragraph{Generation model post-training.}
We use LTX-2~\citep{hacohen2026ltx} as the policy and retain OmniNFT's reinforcement-learning framework~\citep{zhang2026omninft}, but replace its independent expert rewards with VA-Judger. VA-Judger evaluates pairwise candidates and provides audio, video, and shared cross-modal rewards derived from its five rubric dimensions. We freeze the LTX-2 backbone and optimize only LoRA parameters~\citep{hu2021lora}. Full reward construction and training details are provided in Appendix~\ref{app:generator-details}.

\subsection{Reward Model Evaluation}
\label{sec:reward-model-results}
\input{table_judger_accuracy}
We evaluate seven representative automatic metrics spanning unimodal quality, text consistency, audio-video consistency, synchronization, and aggregate quality~\citep{liu2026improving,tjandra2025meta,hessel2021clipscore,girdhar2023imagebind,iashin2024synchformer,liu2026javisdit}. Each metric selects the clip with the better scalar score, with an exact tie receiving half credit. Table~\ref{tab:judger-accuracy} shows limited agreement with human preference: overall micro accuracy ranges from 50.35\% to 57.04\%, while VideoAlign and Audiobox Aesthetics drop to 48.40\% and 44.00\% out of domain.

Combining all seven metrics does not close the gap. Label-free Z-score soft voting improves overall accuracy to 58.70\%, but reaches only 52.40\% out of domain, while hard voting reaches 50.00\%. Separate metrics observe only partial aspects of the joint experience, and embedding similarity or temporal offset can overlook local defects and semantic mismatch. Appendix~\ref{app:metric-fusion} gives the ensemble definitions, and Appendix Tables~\ref{tab:metric-definitions}--\ref{tab:all-sync-metric-acc} define the metrics and report the complete results.

VA-Judger reaches 68.43\% overall Total Acc, improving over Qwen3-Omni Instruct with CoT by 10.60 percentage points. Total Acc counts invalid final choices as incorrect, while Parsed Acc uses only valid responses.

\subsection{Post-training LTX-2 with VA-Judger}
\label{sec:generator-posttraining}

We evaluate all models on 200 uniformly sampled JavisBench prompts~\citep{liu2026javisdit}. Appendix Table~\ref{tab:javisbench-subset-distribution} reports the subset distribution. Higher is better except for DeSync.

\FloatBarrier
\begin{table}[t!]
    \centering
    \caption{Results on the randomly sampled 200-prompt JavisBench subset. LTX-2 + Qwen3-Omni-RL uses the untuned Qwen3-Omni as the reward model under the same post-training configuration as LTX-2 + VA-Judger (Appendix~\ref{app:generator-details}). Best results are highlighted in green and second-best results are underlined. Higher is better unless marked with $\downarrow$.}
    \label{tab:javisbench-results}
    \scriptsize
    \setlength{\tabcolsep}{3.5pt}
    \renewcommand{\arraystretch}{0.96}
    \resizebox{\textwidth}{!}{%
    \begin{tabular}{@{}lrrrrrrrrrr@{}}
        \toprule
        & \multicolumn{2}{c}{Video Quality}
        & \multicolumn{5}{c}{Audiobox Aesthetics}
        & \multicolumn{3}{c}{Cross-Modal Alignment} \\
        \cmidrule(lr){2-3}\cmidrule(lr){4-8}\cmidrule(lr){9-11}
        Model
            & VQ $\uparrow$ & MQ $\uparrow$
            & AQ $\uparrow$ & AB-CE $\uparrow$ & AB-CU $\uparrow$ & AB-PC $\uparrow$ & AB-PQ $\uparrow$
            & TV-Align $\uparrow$ & TA-Align $\uparrow$ & ViCLIP $\uparrow$ \\
        \midrule
        LTX-2
            & 2.248 & 0.697 & 4.767
            & 3.957 & 5.904 & 3.041 & 6.166
            & \underline{0.303} & 0.105 & \underline{0.232} \\
        LTX-2 + Qwen3-Omni-RL
            & 2.908 & 0.635 & 4.900
            & 4.214 & 6.200 & 2.923 & 6.398
            & 0.272 & \underline{0.143} & 0.224 \\
        LTX-2 + OmniNFT
            & \underline{3.727} & \underline{0.947} & \underline{5.399}
            & \underline{4.745} & \cellcolor{bestresult}6.797 & \underline{3.150} & \underline{6.905}
            & 0.279 & 0.133 & 0.219 \\
        LTX-2 + VA-Judger
            & \cellcolor{bestresult}3.942 & \cellcolor{bestresult}1.183 & \cellcolor{bestresult}5.610
            & \cellcolor{bestresult}5.136 & \underline{6.766} & \cellcolor{bestresult}3.606 & \cellcolor{bestresult}6.932
            & \cellcolor{bestresult}0.310 & \cellcolor{bestresult}0.180 & \cellcolor{bestresult}0.242 \\
        \bottomrule
    \end{tabular}}

    \vspace{3pt}
    \resizebox{\textwidth}{!}{%
    \begin{tabular}{@{}lrrrrrrrrrr@{}}
        \toprule
        & \multicolumn{4}{c}{Text Consistency}
        & \multicolumn{6}{c}{AV Consistency and Synchrony} \\
        \cmidrule(lr){2-5}\cmidrule(lr){6-11}
        Model
            & TV-IB $\uparrow$ & TA-IB $\uparrow$ & CLIP $\uparrow$ & CLAP $\uparrow$
            & AV-IB $\uparrow$ & CAVP $\uparrow$ & AV-Align $\uparrow$
            & AVHScore $\uparrow$ & DeSync $\downarrow$ & JavisScore $\uparrow$ \\
        \midrule
        LTX-2
            & 0.355 & 0.102 & 0.303 & 0.304
            & 0.091 & 0.786 & 0.192
            & 0.091 & 0.430 & 0.074 \\
        LTX-2 + Qwen3-Omni-RL
            & \underline{0.370} & \underline{0.140} & \underline{0.317} & 0.354
            & 0.142 & 0.799 & 0.200
            & 0.142 & \underline{0.335} & 0.122 \\
        LTX-2 + OmniNFT
            & 0.336 & 0.134 & 0.298 & \underline{0.394}
            & \underline{0.154} & \cellcolor{bestresult}0.800 & \underline{0.208}
            & \underline{0.146} & \cellcolor{bestresult}0.226 & \underline{0.122} \\
        LTX-2 + VA-Judger
            & \cellcolor{bestresult}0.376 & \cellcolor{bestresult}0.179 & \cellcolor{bestresult}0.326 & \cellcolor{bestresult}0.425
            & \cellcolor{bestresult}0.265 & \underline{0.799} & \cellcolor{bestresult}0.228
            & \cellcolor{bestresult}0.261 & 0.592 & \cellcolor{bestresult}0.230 \\
        \bottomrule
    \end{tabular}}
\end{table}

\paragraph{Quantitative results.}
VA-Judger ranks first on 17 of the 20 reported metrics and receives the highest human preference. Relative to LTX-2, visual, motion, and audio quality rise from 2.248, 0.697, and 4.767 to 3.942, 1.183, and 5.610, while JavisScore and AVHScore rise from 0.074 and 0.091 to 0.230 and 0.261. OmniNFT obtains the lowest DeSync value, but DeSync is explicitly included in its post-training reward. This advantage likely reflects direct optimization of that metric and is consistent with metric-specific reward hacking rather than uniformly better video-audio quality. The broad improvement of VA-Judger is consistent with learning a unified reward from holistic human comparisons rather than optimizing separate expert metrics~\citep{zhang2026omninft}.

\begin{figure}[h!]
    \centering
    \includegraphics[width=\linewidth]{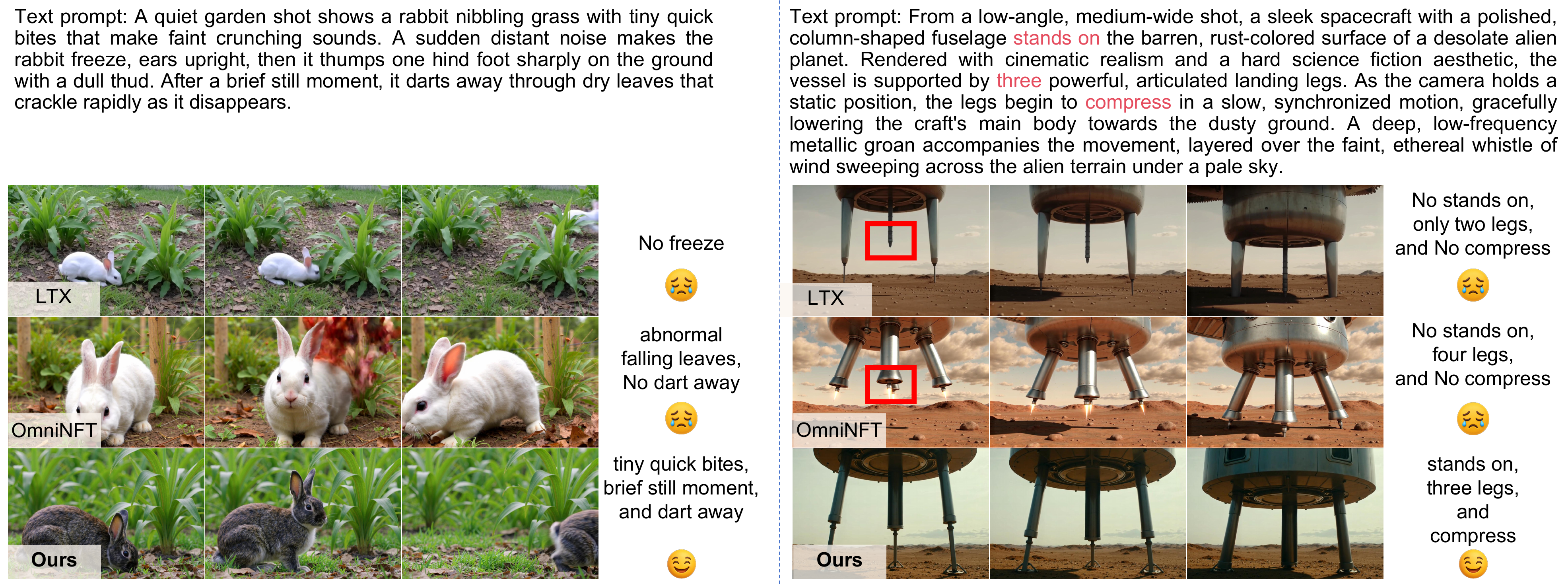}
    \caption{Qualitative comparisons of LTX-2, OmniNFT, and our VA-Judger post-trained model. VA-Judger better follows the requested temporal actions in both examples.}
    \label{fig:ltx2-qualitative}
\end{figure}

\paragraph{Qualitative results.}
Figure~\ref{fig:ltx2-qualitative} shows that the VA-Judger post-trained model better follows the requested temporal actions. In the rabbit example, the baselines omit either the requested freeze-and-dart motion or the intended temporal sequence, whereas our result preserves the brief pause followed by rapid movement. In the spacecraft example, our result maintains the requested three-leg structure and depicts the synchronized compression, while the baselines exhibit incorrect support geometry or fail to show the specified motion.

\begin{figure}[h!]
    \centering
    \includegraphics[width=0.72\linewidth]{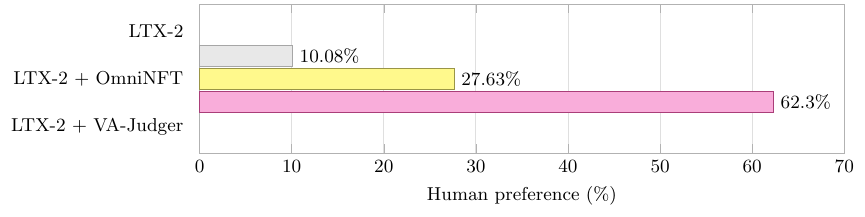}
    \caption{Human preference rates for 200 prompts, aggregated over 4,000 three-way selections by 20 participants, among LTX-2, LTX-2 post-trained with OmniNFT, and LTX-2 post-trained with VA-Judger as the learned reward model.}
    \label{fig:human-preference}
\end{figure}

\paragraph{Human evaluation.}
Across 4,000 selections from 200 three-way comparisons judged by 20 adults, VA-Judger receives 62.30\% of preferences, compared with 27.63\% for OmniNFT and 10.08\% for LTX-2, as shown in Figure~\ref{fig:human-preference}.

\FloatBarrier
\section{Conclusion}

We introduced VA-Judger, a human-aligned omni-reward model for joint video-audio generation, together with VAPref-10K and VA-Judger-Bench. VA-Judger combines easy-to-hard supervised fine-tuning with human-matched hard examples and dimension-wise reinforcement learning to learn final preferences and structured dimension-level comparisons. Unlike separate expert metrics that optimize isolated properties, its unified reward jointly evaluates visual quality, audio quality, semantic fidelity, and cross-modal coherence. On VA-Judger-Bench, it improves preference prediction over existing omni-models and automatic metrics across both in-domain and out-of-domain evaluations. As a post-training reward for LTX-2, it leads on 17 of 20 automatic metrics and receives 62.30\% human preference. Together, these results establish holistic human preference supervision as an effective foundation for improving joint video-audio generation.

\bibliography{VA-Judger_arxiv}
\bibliographystyle{iclr2027_conference}

\clearpage
\appendix
\phantomsection
\label{app:appendix}
\section*{Appendix}

\section{Qualitative Example of a VA-Judger Judgment}
\label{app:full-case}

\begin{center}
    \includegraphics[width=\linewidth,height=0.87\textheight,trim={20bp 39bp 17bp 13bp},clip]{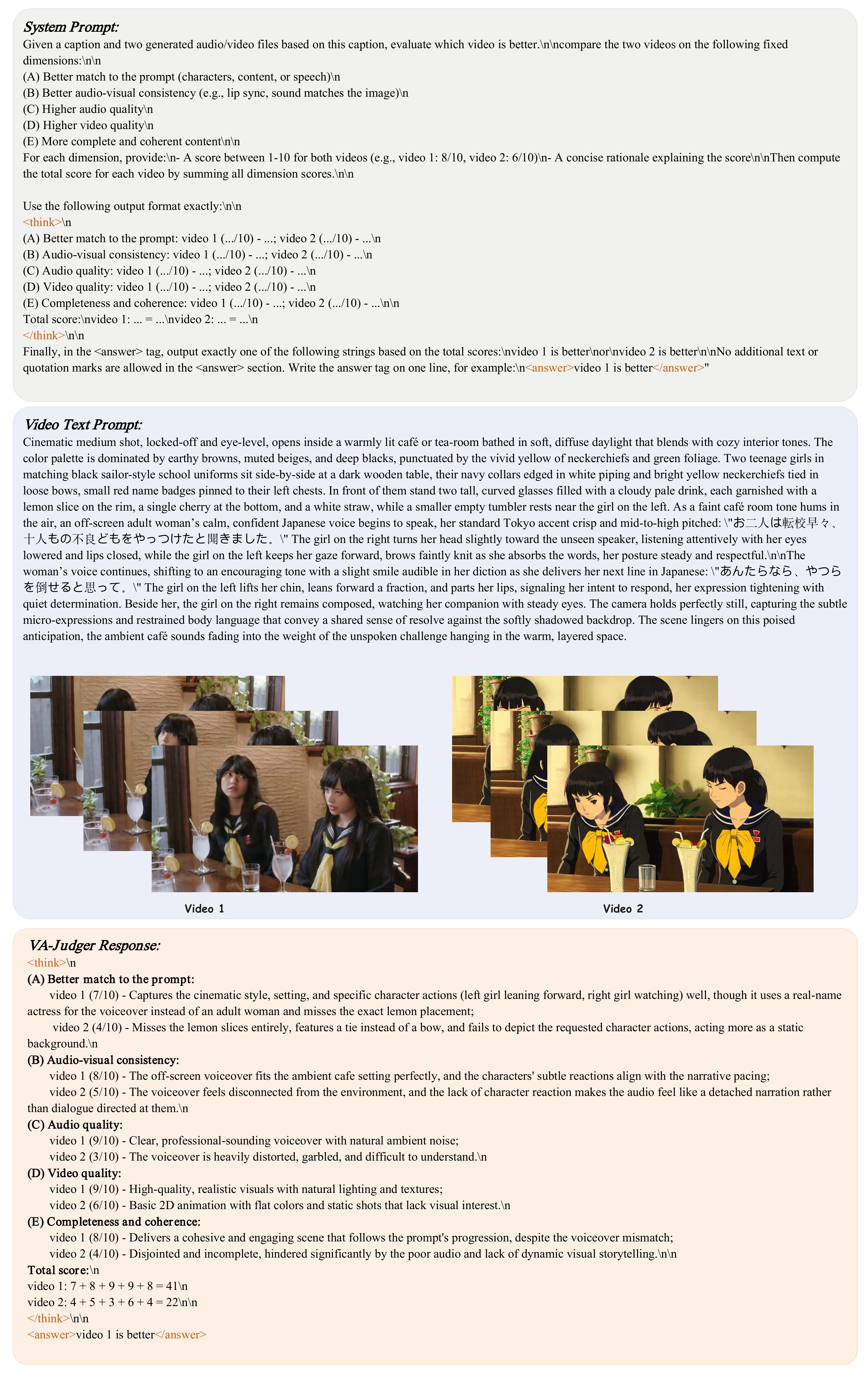}
    \refstepcounter{figure}
    \label{fig:full-case}
    \vspace{2pt}

    \begin{minipage}{0.98\linewidth}
    \small Figure~\thefigure: A complete VA-Judger input and response. The system prompt defines dimensions A to E, and the response compares both clips on every dimension before producing the final preference.
    \end{minipage}
\end{center}
\clearpage

\section{VAPref-10K Construction and Final Pair Distribution}
\label{app:vapref-construction}

\subsection{Construction Pipeline}

\begin{figure}[h!]
    \centering
    \includegraphics[width=\linewidth,]{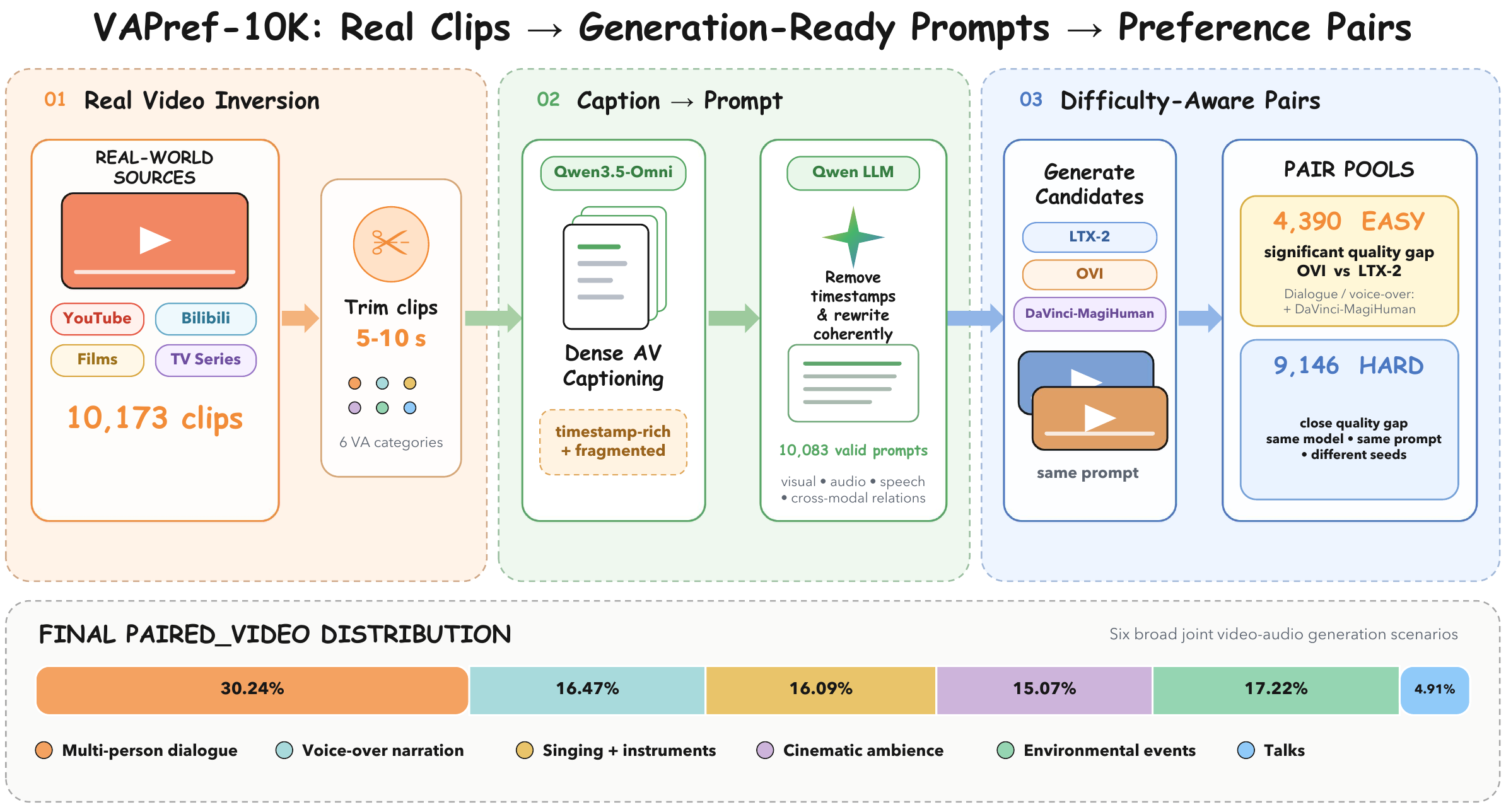}
    \vspace{-0.7cm}
    \caption{Overview of VAPref-10K construction and the distribution of paired generated videos retained after filtering and pair construction. Real video-audio clips are captioned by Qwen3.5-Omni, rewritten into text prompts by an LLM, and passed to LTX-2, Ovi, and daVinci-MagiHuman to generate candidate clips. A prompt may yield multiple generated clips and therefore contribute more than one within-model comparison. The diagram reports 4,390 easy cold-start pairs and 9,146 human-annotated raw hard pairs. The separately constructed 1,150-pair benchmark is not included in the diagram.}
    \label{fig:data_pipeline}
\end{figure}
\FloatBarrier

\subsection{Source Video Distribution}
\label{app:video-category-distribution}

\begin{figure}[h!]
    \centering
    \includegraphics[width=0.92\linewidth]{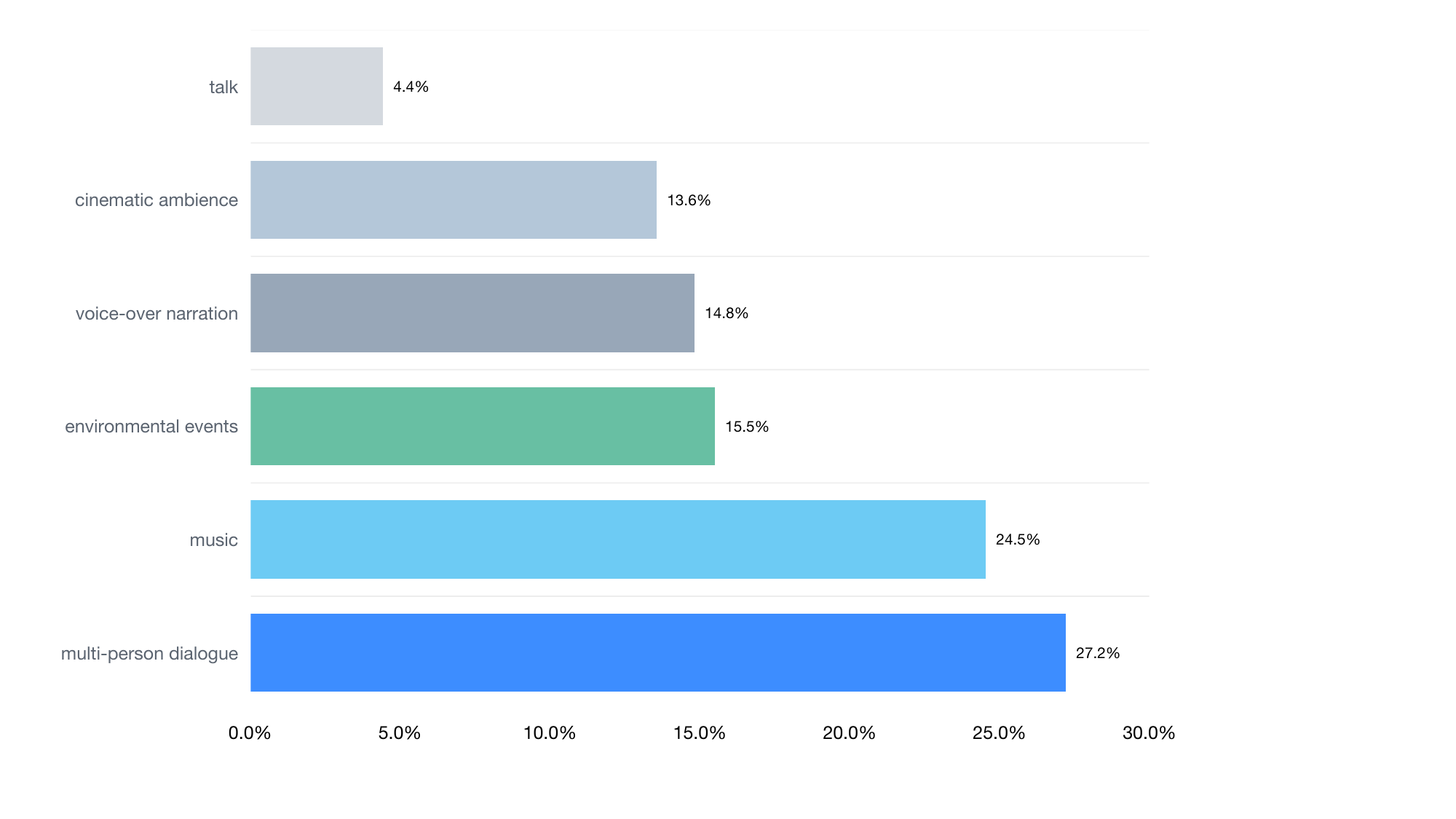}
    \caption{Distribution of the source videos across the six categories in VAPref-10K.}
    \label{fig:video-category-distribution}
\end{figure}
\FloatBarrier

\section{JavisBench Evaluation Subset Distribution}
\label{app:javisbench-subset-distribution}

Table~\ref{tab:javisbench-subset-distribution} compares the category distribution of the full 10,140 prompt JavisBench pool with the 200 prompt subset used for generation model evaluation. The subset is sampled uniformly at random without category-based filtering or stratification. Sound Type and Spatial Composition allow multiple labels, so their percentages do not sum to 100\%.

\begin{table}[t!]
    \centering
    \caption{Category distributions of the full JavisBench pool and the random evaluation subset. Each cell reports the number and percentage of examples.}
    \label{tab:javisbench-subset-distribution}
    \scriptsize
    \setlength{\tabcolsep}{5.5pt}
    \renewcommand{\arraystretch}{1.02}
    \begin{tabular}{@{}llrr@{}}
        \toprule
        Dimension & Category & Full pool (10,140) & Random subset (200) \\
        \midrule
        Event Scenario & Living Scenario       & 4,029 (39.7\%) & 73 (36.5\%) \\
                       & Natural Scenario      & 2,452 (24.2\%) & 37 (18.5\%) \\
                       & Urban Scenario        & 2,140 (21.1\%) & 66 (33.0\%) \\
                       & Industrial Scenario   & 1,090 (10.7\%) & 20 (10.0\%) \\
                       & Virtual Scenario      &   429 (4.2\%)  & 4 (2.0\%) \\
        \addlinespace
        Visual Style & Camera Shooting       & 8,586 (84.7\%) & 174 (87.0\%) \\
                     & 2D Animation          &   915 (9.0\%)  & 17 (8.5\%) \\
                     & 3D Animation          &   639 (6.3\%)  & 9 (4.5\%) \\
        \addlinespace
        Sound Type & Musical Sounds        & 5,065 (50.0\%) & 73 (36.5\%) \\
                   & Ambient Sounds        & 4,183 (41.3\%) & 126 (63.0\%) \\
                   & Biological Sounds     & 2,215 (21.8\%) & 59 (29.5\%) \\
                   & Mechanical Sounds     & 1,940 (19.1\%) & 49 (24.5\%) \\
                   & Speech Sounds         & 1,090 (10.7\%) & 29 (14.5\%) \\
        \addlinespace
        Spatial Composition & Multiple Subjects     & 7,588 (74.8\%) & 167 (83.5\%) \\
                            & Single Subject        & 2,502 (24.7\%) & 33 (16.5\%) \\
                            & Off-screen Sound      & 2,165 (21.4\%) & 59 (29.5\%) \\
        \addlinespace
        Temporal Composition & Sequential Events     & 3,698 (36.5\%) & 46 (23.0\%) \\
                             & Simultaneous Events   & 3,306 (32.6\%) & 99 (49.5\%) \\
                             & Single Event          & 3,136 (30.9\%) & 55 (27.5\%) \\
        \bottomrule
    \end{tabular}
\end{table}
\FloatBarrier

\section{Implementation Details}

\subsection{Reward Model Training and Evaluation}
\label{app:reward-eval-details}

\paragraph{Supervised fine-tuning.}
Stage 1 initializes from Qwen3-Omni-30B-A3B-Instruct and trains on easy pairs for 400 optimizer updates. Stage 2 continues from the Stage 1 checkpoint and trains on hard pairs for another 1,000 updates. In both stages, we update the language backbone while freezing the visual encoder and aligner. Training uses eight GPUs, BF16, DeepSpeed ZeRO 3, a peak learning rate of $5\times10^{-6}$, a warmup ratio of 0.05, a maximum sequence length of 24,576, and an effective batch size of 128.

\paragraph{Dimension-wise GRPO.}
This stage starts from the hard SFT checkpoint and trains for 600 optimizer updates on 9,136 human-annotated hard pairs after excluding 10 only-``Other'' annotations. We freeze the visual and audio encoders and update the language thinker using six training GPUs, while two GPUs host rollout servers. We sample eight responses per pair with temperature 1.0 and optimize with Adafactor at a learning rate of $1\times10^{-6}$. The answer and dimension rewards receive equal weight.

\paragraph{Evaluation.}
We use greedy decoding with vLLM. Each process receives the training comparison rubric. We extract the final choice from the \texttt{<answer>} tag; a response without a valid choice is incorrect for Total Acc and excluded from Parsed Acc.

\subsection{LTX-2 Post-training}
\label{app:generator-details}

We initialize the policy from the 19B LTX-2 checkpoint and keep the backbone frozen. LoRA adapters with rank 32 and scaling factor 64 are inserted into the video and audio attention layers, feed-forward layers, and bidirectional cross-modal attention layers. The current and old policies use separate adapters, while disabling the adapters recovers the fixed reference policy.

For each prompt, the old policy generates eight video-audio candidates, which form 28 unique pairs. VA-Judger evaluates every pair using the five rubric dimensions shown in Appendix Figure~\ref{fig:full-case}. We average the scores obtained by each candidate across its seven comparisons. Dimension C provides the audio reward, dimension D provides the video reward, and the average of dimensions A, B, and E provides the shared cross-modal reward. Each reward is normalized within the eight-candidate group. The audio and video rewards are routed to their corresponding generation branches, while the shared reward updates both branches. Gemini 3.1 Pro~\citep{google2026gemini31pro} is used only for offline data construction, while online post-training uses the locally deployed VA-Judger for efficient batched reward computation.

Following OmniNFT~\citep{zhang2026omninft}, we use its 20K-prompt VGGSound split and train for 400 optimizer updates. Candidates contain 121 frames at $512\times768$ resolution and 24 FPS, and rollouts use 20 denoising steps. We optimize the LoRA parameters with AdamW at a learning rate of $3\times10^{-5}$ in BF16. Training runs with FSDP on seven GPUs, while one additional GPU hosts a VA-Judger inference server. All generation-model baselines use the same prompts, generation settings, and evaluation subset.

\paragraph{Qwen3-Omni-RL baseline.}
Qwen3-Omni-RL in Table~\ref{tab:javisbench-results} isolates the contribution of reward-model training by replacing VA-Judger with the untuned Qwen3-Omni-30B-A3B-Instruct. It uses the same rubric, pairwise aggregation, LoRA configuration, prompts, and post-training settings as LTX-2 + VA-Judger.

\section{Full Metric Alignment Results}
\label{app:all-metric-human-acc}

We convert each scalar metric into a pairwise prediction using its preferred direction and compare the prediction with the human label. Every metric is evaluated on all 1,150 benchmark pairs, comprising 400 easy, 250 in-domain, and 500 out-of-domain comparisons. Exact score ties receive half credit. Metrics selected as representatives in Table~\ref{tab:judger-accuracy} are marked with $\ast$.

\begin{table}[t!]
    \centering
    \caption{Definitions and implementation details for less standard automatic metrics. All implementations use the JavisBench evaluation pipeline~\citep{liu2026javisdit}; $\uparrow$ and $\downarrow$ indicate the preferred raw-score direction before conversion to pairwise accuracy.}
    \label{tab:metric-definitions}
    \scriptsize
    \setlength{\tabcolsep}{3.5pt}
    \renewcommand{\arraystretch}{1.06}
    \begin{tabular}{@{}p{0.12\textwidth}p{0.25\textwidth}p{0.12\textwidth}cp{0.36\textwidth}@{}}
        \toprule
        Metric & Definition & Input & Dir. & Preprocessing and source \\
        \midrule
        VideoPhy2 & VideoPhy-2-AutoEval for semantic adherence and physical commonsense~\citep{bansal2026videophy} & Text, video & $\uparrow$ & Decode and sample the generated clip following the official evaluator through the JavisBench wrapper. \\
        NISQA & Non-Intrusive Speech Quality Assessment, predicting perceptual speech quality~\citep{mittag2021nisqa} & Audio & $\uparrow$ & Extract audio and apply the model-native waveform preprocessing and sampling rate. \\
        DNSMOS & Deep Noise Suppression Mean Opinion Score, using the overall P.835 quality prediction~\citep{reddy2022dnsmos} & Audio & $\uparrow$ & Extract audio, convert it to the model-required mono waveform, and apply model-native resampling. \\
        ViCLIP & Video-text contrastive similarity from InternVid~\citep{wang2024internvid} & Text, video & $\uparrow$ & Sample video frames with the JavisBench wrapper and compute cosine similarity between video and text embeddings. \\
        SyncNet & Audio-mouth synchronization network~\citep{chung2016out} & Audio, face video & Mixed & Use the JavisBench face-track and audio preprocessing. Confidence is higher-better; minimum embedding distance and absolute offset are lower-better. \\
        \bottomrule
    \end{tabular}
\end{table}

\subsection{Label-Free Metric Fusion}
\label{app:metric-fusion}

We examine whether combining the seven representative automatic metrics provides a stronger proxy for human preference. The ensemble contains VideoAlign, Audiobox Aesthetics, CLIP Score, ImageBind text-audio similarity, ImageBind audio-video similarity, SynchFormer offset, and Javis Score. Both fusion methods assign equal importance to all metrics and do not use human preference labels to learn or tune weights. They are evaluated on the same 1,150 benchmark pairs as the individual metrics.

\paragraph{Equal-weight hard voting.}
For each pair, every metric votes for the clip with the better score. We orient all votes so that $v_m=+1$ selects the first clip and $v_m=-1$ selects the second clip. Higher values are preferred for all metrics except SynchFormer offset, for which a lower value is preferred. An exact score tie contributes no vote. The ensemble prediction is
\begin{equation}
    \hat y_{\mathrm{hard}} = \operatorname{sign}\!\left(\sum_{m=1}^{7} v_m\right).
\end{equation}
Hard voting obtains 64.50\%, 52.00\%, and 50.00\% accuracy on the easy, in-domain, and out-of-domain splits, respectively. The resulting micro accuracy over all 1,150 benchmark pairs is 55.48\%.

\paragraph{Equal-weight Z-score soft voting.}
Hard voting ignores the magnitude of each score difference. We therefore also standardize the scores of each metric within each evaluation split,
\begin{equation}
    z_m(x)=\frac{s_m(x)-\mu_m}{\sigma_m},
    \qquad
    S(x)=\frac{1}{7}\sum_{m=1}^{7}\tilde z_m(x),
\end{equation}
where $\mu_m$ and $\sigma_m$ are computed from all clip scores for metric $m$ in the corresponding split. We set $\tilde z_m(x)=-z_m(x)$ for SynchFormer offset and $\tilde z_m(x)=z_m(x)$ for the other six metrics so that a higher value is always preferred. The clip with the larger $S(x)$ is selected. This method obtains 67.00\%, 58.00\%, and 52.40\% accuracy on the three splits, with an overall micro accuracy of 58.70\%. Compared with Javis Score, the strongest of the seven representative individual metrics, Z-score soft voting improves easy, in-domain, and overall accuracy by 7.00, 2.40, and 1.66 percentage points, respectively, but decreases out-of-domain accuracy by 3.00 points. This result shows that the metrics contain complementary information, while their label-free combination remains insufficiently robust under generation model distribution shift.

\begin{table}[t!]
    \centering
    \caption{Human preference accuracy (\%) of video-quality metrics. All raw metric scores are higher-better before pairwise conversion.}
    \label{tab:all-video-metric-acc}
    \scriptsize
    \setlength{\tabcolsep}{5.2pt}
    \renewcommand{\arraystretch}{1.08}
    \begin{tabular}{@{}lrrrr@{}}
        \toprule
        Metric & Easy & In-domain & Out-of-domain & Overall (micro) \\
        \midrule
        VideoAlign Overall$^{\ast}$ & 62.25 & 53.20 & 48.40 & 54.26 \\
        VideoPhy2 & 52.75 & 54.00 & 50.20 & 51.91 \\
        Motion Quality & 60.00 & 51.20 & 50.00 & 53.74 \\
        Visual Quality & 62.50 & 52.40 & 42.80 & 51.74 \\
        HPSv3 & 56.50 & 52.00 & 50.20 & 52.78 \\
        Video Aesthetic & 57.75 & 50.00 & 47.40 & 51.57 \\
        Video Technical/Aesthetic & 57.50 & 51.20 & 42.80 & 49.74 \\
        Video Technical/Overall & 59.25 & 52.80 & 42.20 & 50.43 \\
        Video Technical/Technical & 54.25 & 54.00 & 47.20 & 51.13 \\
        \bottomrule
    \end{tabular}
\end{table}

\begin{table}[t!]
    \centering
    \caption{Human preference accuracy (\%) of audio-quality metrics. All raw metric scores are higher-better before pairwise conversion.}
    \label{tab:all-audio-metric-acc}
    \scriptsize
    \setlength{\tabcolsep}{5.2pt}
    \renewcommand{\arraystretch}{1.08}
    \begin{tabular}{@{}lrrrr@{}}
        \toprule
        Metric & Easy & In-domain & Out-of-domain & Overall (micro) \\
        \midrule
        Audiobox Aesthetics$^{\ast}$ & 58.50 & 50.00 & 44.00 & 50.35 \\
        Audiobox Aesthetics CE & 61.50 & 50.00 & 42.20 & 50.61 \\
        Audiobox Aesthetics CU & 62.25 & 48.00 & 38.60 & 48.87 \\
        Audiobox Aesthetics PC & 45.25 & 54.40 & 53.60 & 50.87 \\
        Audiobox Aesthetics PQ & 64.25 & 45.60 & 36.20 & 48.00 \\
        NISQA & 53.75 & 51.60 & 43.80 & 48.96 \\
        DNSMOS & 47.25 & 52.00 & 40.40 & 45.30 \\
        \bottomrule
    \end{tabular}
\end{table}

\begin{table}[t!]
    \centering
    \caption{Human preference accuracy (\%) of text-video and text-audio consistency metrics. All raw similarity scores are higher-better before pairwise conversion.}
    \label{tab:all-text-metric-acc}
    \scriptsize
    \setlength{\tabcolsep}{5.2pt}
    \renewcommand{\arraystretch}{1.08}
    \begin{tabular}{@{}lrrrr@{}}
        \toprule
        Metric & Easy & In-domain & Out-of-domain & Overall (micro) \\
        \midrule
        \multicolumn{5}{@{}l}{\textit{Text-video consistency}} \\
        CLIP Score$^{\ast}$ & 55.75 & 54.80 & 53.80 & 54.70 \\
        ImageBind T-V & 47.25 & 53.20 & 51.00 & 50.17 \\
        ViCLIP & 53.75 & 43.60 & 49.60 & 49.74 \\
        \midrule
        \multicolumn{5}{@{}l}{\textit{Text-audio consistency}} \\
        ImageBind T-A$^{\ast}$ & 58.50 & 52.80 & 51.60 & 54.26 \\
        CLAP Score & 51.25 & 52.00 & 66.20 & 57.91 \\
        \bottomrule
    \end{tabular}
\end{table}

\begin{table}[t!]
    \centering
    \caption{Human preference accuracy (\%) of audio-video semantic-consistency metrics. All raw scores are higher-better. Audio-Video Alignment, AVHScore, and ImageBind A-V are ImageBind-based metrics but differ in preprocessing and feature aggregation.}
    \label{tab:all-av-semantic-metric-acc}
    \scriptsize
    \setlength{\tabcolsep}{5.2pt}
    \renewcommand{\arraystretch}{1.08}
    \begin{tabular}{@{}lrrrr@{}}
        \toprule
        Metric & Easy & In-domain & Out-of-domain & Overall (micro) \\
        \midrule
        Audio-Video Alignment$^{\ast}$ & 61.25 & 55.60 & 51.80 & 55.91 \\
        AVH Score & 61.75 & 58.80 & 54.40 & 57.91 \\
        ImageBind A-V & 61.25 & 58.80 & 55.20 & 58.09 \\
        
        \bottomrule
    \end{tabular}
\end{table}

\begin{table}[t!]
    \centering
    \caption{Human preference accuracy (\%) of synchronization and aggregate audio-video metrics. Preferred raw-score directions are repeated beside each metric.}
    \label{tab:all-sync-metric-acc}
    \scriptsize
    \setlength{\tabcolsep}{4.2pt}
    \renewcommand{\arraystretch}{1.08}
    \begin{tabular}{@{}lrrrr@{}}
        \toprule
        Metric & Easy & In-domain & Out-of-domain & Overall (micro) \\
        \midrule
        SyncNet Confidence $\uparrow$ & 64.50 & 50.80 & 40.20 & 50.96 \\
        SyncNet Min Distance $\downarrow$ & 54.50 & 59.20 & 47.00 & 52.26 \\
        SyncNet $|\mathrm{Offset}|$ (frames) $\downarrow$ & 65.25 & 48.00 & 42.00 & 51.39 \\
        SynchFormer $|\mathrm{Argmax\ Offset}|$ (s) $\downarrow$ & 63.75 & 47.20 & 50.60 & 54.43 \\
        SynchFormer $|\mathrm{Offset}|$ (s)$^{\ast}$ $\downarrow$ & 61.50 & 46.80 & 48.20 & 52.52 \\
        Desync $\downarrow$ & 66.25 & 49.20 & 52.00 & 56.35 \\
        Lip-sync Confidence $\uparrow$ & 63.25 & 53.60 & 42.80 & 52.26 \\
        Javis Score$^{\ast}$ $\uparrow$ & 60.00 & 55.60 & 55.40 & 57.04 \\
        \bottomrule
    \end{tabular}
\end{table}
\FloatBarrier

\end{document}

%% file: math_commands.tex
\usepackage{amsmath,amsfonts,bm}

\def\eqref#1{equation~\ref{#1}}
\def\1{\bm{1}}

\DeclareMathAlphabet{\mathsfit}{\encodingdefault}{\sfdefault}{m}{sl}
\SetMathAlphabet{\mathsfit}{bold}{\encodingdefault}{\sfdefault}{bx}{n}

%% file: table_judger_accuracy.tex
\begin{table}[t!]
    \centering
    \caption{Accuracy (\%) against human pairwise preferences on 400 easy, 250 in-domain, and 500 out-of-domain pairs. Reward models report Total Acc / Parsed Acc. Overall is the micro accuracy over all 1,150 pairs.}
    \label{tab:judger-accuracy}
    \scriptsize
    \setlength{\tabcolsep}{4.2pt}
    \renewcommand{\arraystretch}{1.00}
    \begin{tabular}{@{}lcccc@{}}
        \toprule
        Model
            & Easy
            & In-domain
            & Out-of-domain
            & Overall (micro) \\
        \midrule
        \multicolumn{5}{@{}l}{\textit{Automatic evaluation metrics}} \\
        Video quality: VideoAlign
            & 62.25 & 53.20 & 48.40 & 54.26 \\
        Audio quality: Audiobox Aesthetics
            & 58.50 & 50.00 & 44.00 & 50.35 \\
        Text-video: CLIP Score
            & 55.75 & 54.80 & 53.80 & 54.70 \\
        Text-audio: ImageBind T-A
            & 58.50 & 52.80 & 51.60 & 54.26 \\
        Audio-video: ImageBind
            & 61.25 & 55.60 & 51.80 & 55.91 \\
        Synchronization: SynchFormer
            & 61.50 & 46.80 & 48.20 & 52.52 \\
        Overall: Javis Score
            & 60.00 & 55.60 & \textbf{55.40} & 57.04 \\
        Metric ensemble: Hard voting
            & 64.50 & 52.00 & 50.00 & 55.48 \\
        Metric ensemble: Z-score soft voting
            & \textbf{67.00} & \textbf{58.00} & 52.40 & \textbf{58.70} \\
        \midrule
        \multicolumn{5}{@{}l}{\textit{Reward models}} \\
        Qwen3-Omni Captioner (No CoT)
            & 57.00 / 57.00 & 54.00 / 54.00 & 56.20 / 56.20 & 56.00 / 56.00 \\
        Qwen3-Omni Captioner (CoT)
            & 52.50 / 57.69 & 49.60 / 54.87 & 46.60 / 56.97 & 49.30 / 56.76 \\
        \midrule
        Qwen3-Omni Instruct NoCoT
            & 60.75 / 60.75 & 54.00 / 54.00 & 54.60 / 54.60 & 56.61 / 56.61 \\
        Qwen3-Omni Instruct CoT
            & 63.25 / 64.71 & 54.80 / 55.92 & 55.00 / 55.33 & 57.83 / 58.69 \\
        \midrule
        \textbf{VA-Judger (Ours)}
            & \textbf{76.25} / 76.25 & \textbf{66.00} / 66.00 & \textbf{63.40} / 63.40 & \textbf{68.43} / 68.43 \\
        \bottomrule
    \end{tabular}
\end{table}